\documentclass[twoside,leqno,twocolumn]{article}
\usepackage{ltexpprt}

\renewcommand{\thefootnote}{\arabic{footnote}}
\usepackage{url}
\usepackage{array}
\usepackage{fontawesome}
\usepackage{amsmath,amssymb}
\usepackage{multirow}
\usepackage{hhline}
\usepackage{tabulary}
\usepackage[pdftex]{graphicx}
\usepackage{lipsum}
\usepackage[a4paper,margin=1in]{geometry}
\usepackage{fancyhdr}
\begin{document}

\title{A Practitioners' Guide to Transfer Learning for Text Classification using Convolutional Neural Networks}
\author{Tushar Semwal\\
Indian Institute of Technology Guwahati\\
t.semwal@iitg.ac.in\\
\\
\and
Gaurav Mathur \\
Samsung R\&D Institute-Bangalore\\
gaurav.m4@samsung.com\\
\and
Promod Yenigalla\\
Samsung R\&D Institute-Bangalore\\
promod.y@samsung.com\\
\and
Shivashankar B. Nair\\
Indian Institute of Technology Guwahati\\
sbnair@iitg.ac.in\\
}
\date{}

\maketitle
\thispagestyle{fancy}

% Copyright Statement
% When submitting your final paper to a SIAM proceedings, it is requested that you include 
% the appropriate copyright in the footer of the paper.  The copyright added should be 
% consistent with the copyright selected on the copyright form submitted with the paper.
% Please note that "20XX" should be changed to the year of the meeting.

% Default Copyright Statement
\fancyfoot[R]{\footnotesize{\textbf{Copyright \textcopyright\ 20xx by SIAM\\
Unauthorized reproduction of this article is prohibited}}}

% Depending on which copyright you agree to when you sign the copyright form, the copyright 
% can be changed to one of the following after commenting out the default copyright statement
% above.

%\fancyfoot[R]{\footnotesize{\textbf{Copyright \textcopyright\ 20XX\\
%Copyright for this paper is retained by authors}}}

%\fancyfoot[R]{\footnotesize{\textbf{Copyright \textcopyright\ 20XX\\
%Copyright retained by principal author's organization}}}

%\pagenumbering{arabic}
%\setcounter{page}{1}%Leave this line commented out.

\begin{abstract} \small\baselineskip=9pt ﻿Transfer Learning (TL) plays a crucial role when a given dataset has insufficient labeled examples to train an accurate model. In such scenarios, the knowledge accumulated within a model pre-trained on a source dataset can be transferred to a target dataset, resulting in the improvement of the target model. Though TL is found to be successful in the realm of image-based applications, its impact and practical use in Natural Language Processing (NLP) applications is still a subject of research. Due to their hierarchical architecture, Deep Neural Networks (DNN) provide flexibility and customization in adjusting their parameters and depth of layers, thereby forming an apt area for exploiting the use of TL. In this paper, we report the results and conclusions obtained from extensive empirical experiments using a Convolutional Neural Network (CNN) and try to uncover thumb rules to ensure a successful \textit{positive} transfer. In addition, we also highlight the flawed means that could lead to a \textit{negative} transfer. We explore the transferability of various layers and describe the effect of varying hyper-parameters on the transfer performance. Also, we present a comparison of accuracy value and model size against state-of-the-art methods. Finally, we derive inferences from the empirical results and provide best practices to achieve a successful positive transfer.\end{abstract}

\section{Introduction}
Transfer Learning (TL) is a branch of Machine Learning (ML) which leverages the knowledge stored within a \textit{source} domain and provides a means to transfer the same to a \textit{target} domain where a domain could mean datasets, tasks, games, etc. It is based on the fact that features that have been learned, for instance, to classify ripe and non ripe apples can also be used to classify pears or peaches - fruits of the same family.  TL becomes indispensable in scenarios where a given task (referred to as the  \textit{target} task) does not have enough data required to train and accurate model. Under such situations, the knowledge stored within an off-the-shelf model which is trained on a source domain can be transferred to the target domain. 
TL has been used effectively in the domain of image processing. Of late, it has shown great promise when applied to several applications in the domain of Natural Language Processing (NLP).  

Deep Neural Networks (DNN) have long been a subject of interest especially in the context of TL, more specifically in the domain of image processing.  Krizhevsky and Lee \cite{krizhevsky2012imagenet,lee:2009} demonstrate how low-level neural layers in such networks can be successfully transferred for different tasks. In \cite{donahue2013decaf,sermanet2014} high-level layers from a DNN trained on source dataset have been transferred to a DNN with a smaller sized target dataset. Transferability of each layer of a DNN has been studied in \cite{yosinski2014transferable} where they have shown that while the lower layers learn features which are general across different tasks, the higher ones reflect the specific nature of the task at hand. For example, in the case of object classification, the hidden parameters of the output regression layer (generally the last layer) are highly specific to the number and type of labels to classify and hence are less likely to be transferable. Due to their stacked architecture, DNN tends to provide flexibility while transferring knowledge stored within their layers, which in turn facilitates TL. 

While TL has produced positive results within the domain of Image Processing, its use in NLP applications still remains an exciting and fairly unexplored area of research. Mou et al. \cite{mou2016HowTA} have provided insights on the transferability of layers of DNN in NLP applications. Using Sentence and Sentence-pair classification as the tasks, they performed experiments for two transfer methods - parameter initialization and multi-task learning. Bowman et al. \cite{bowmanlarge} achieved a remarkable increase in the accuracy of a language understanding task by initializing the parameters through an additional unlabeled dataset. Johnson and Zhang \cite{johnson2015} accomplished a similar feat where they have proposed a semi-supervised framework to increase the text classification accuracy by integrating knowledge from embeddings (word vectors) learned on unlabeled data. Though researchers have applied TL in many forms such as the use of pre-trained word embeddings, transfer of different neural layers, multi-task learning \cite{mou2016HowTA}, semi-supervised learning \cite{bowmanlarge}, domain adaptation \cite{daume2009frustratingly}, etc., a manual (or a practitioners' guide) for applying TL to NLP applications is grossly missing. 

The work presented herein endeavors to provide a set of parameters and rules, that practitioners can directly adapt in their experiments. It is envisaged that this will greatly reduce the efforts invested in finding the best settings and combinations for applying TL. Along with positive transfer scenarios, we also discuss cases and settings which could lead to a negative transfer. In addition, we also compare the accuracy performance and model size of our baseline TL augmented CNN model with other state-of-the-art methods. To the best of our knowledge, this is the first work that presents an analysis and discussions on what to and what not to transfer in actual practical scenarios. Inspired by the empirical analysis of neural models presented in \cite{zhang2015sensitivity,coates2011a,breuel2015a}, we present results obtained from using TL for the task of text classification.

\section{Datasets}
We chose a total of seven datasets for the task of TL and divided them into two sets: a source domain set ($\mathcal{D}_s$) and a target domain set ($\mathcal{D}_t$). Different statistical metrics for the datasets are shown in Table \ref{tab1}. TL comes to the rescue when the target dataset is not large enough for a model to learn a task accurately. Thus all small size datasets form the set $\mathcal{D}_t$ while the remaining large datasets are placed into the set $\mathcal{D}_s$.  As can be seen from the Table \ref{tab1}, the datasets in $\mathcal{D}_s$ have high value of metrics as compared to the ones in $\mathcal{D}_t$ which are smaller in size. A brief description of each of the datasets used follows.

\begin{itemize}
\item \textbf{Amazon (AMZ) reviews:} This includes Amazon product reviews collected by Zhang et al. \cite{zhang2015character} from the Stanford Network Analysis Project. We have used both the polarities (positive and negative, i.e. 2-class) and the 5-class dataset\footnote{\url{https://goo.gl/bm0IkT}}.

%which has 1,800,000 training samples and 200,000 testing samples in each polarity sentiment

\item \textbf{Yelp polarity (YELP-2) reviews:} This dataset is a collection of business reviews and has 280,000 training samples and 19,000 testing samples for  each polarity class\textsuperscript{1} \cite{zhang2015character}. 

\item \textbf{IMDb}: This is a binary dataset with 12,500 multi-sentence movie reviews in each class\footnote{\url{https://goo.gl/NWatud} %{http://ai.stanford.edu/~amaas/data/sentiment/}
} 
\cite{maas2011}. Reviews are in the form of long sentences. In the work presented herein, we have used the first 200 words from each review.

\item \textbf{Movie Reviews (MR):} This is a small sized dataset that contains reviews in the form of sentences. The sentiments of these sentences have been classified as  positive or negative\footnote{\label{note1}{\url{https://www.cs.cornell.edu/people/pabo/movie-review-data/}}} \cite{pang2005lee}.

\item \textbf{Stanford Sentiment Treebank (SST) datasets:} This is a 2-class and 5-class version of the Stanford sentiment analysis dataset. SST-2 has 9,613 samples while SST-5 contains 11,855 reviews\footnote{\label{note2}\url{http://nlp.stanford.edu/sentiment/}}. The SST dataset also contain phrases for each of the sentences. We have used only the phrases for the training phase as practiced in \cite{socher2013recursive,le2014distributed}.

%\item \textbf{Subjectivity dataset:} A 2-class dataset where each sentence is classified either as objective or subjective (Pang and Lee 2014). Though the number of classes are same as in review datasets discussed above, Subjectivity dataset provides an entirely different task to learn.

\end{itemize}

\begin{table}[tbp]
\footnotesize
\centering
\caption{Statistics for the datasets}
\label{tab1}
\begin{tabular}{l||l|l|l|l|l|l}
%\hline
 Dataset & $\mathcal{D}$  & $\mathcal{C}$ & $\mathcal{L}$ & $\mathcal{N}$ & $\mathcal{V}$ & $\mathcal{V}_{pre}$ \\ \hline \hline
AMZ-5 & $\mathcal{D}_s$  & 5 & 84 & 3650000 & 1057296& 120177\\ \hline
AMZ-2 & $\mathcal{D}_s$  & 2  & 82 & 3000000 & 1112820 & 121015 \\ \hline
YELP-2 & $\mathcal{D}_s$ & 2 & 141 & 560000  & 246735 & 77156\\ \hline
IMDb & $\mathcal{D}_{t}$ & 2  & 257 & 25000 & 81321 & 49070\\ \hline
SST-5 & $\mathcal{D}_t$ & 5  & 18  &11855 & 17836 & 16262\\ \hline
MR & $\mathcal{D}_t$ & 2 & 20 & 10662 & 18765 & 16448 \\ \hline
%SUBJ & $\mathcal{D}_t$ & 2 & 23 & 10000&21323 & 17913 \\ \hline
SST-2 & $\mathcal{D}_t$ & 2 & 19 & 9613 &16185 & 14838\\ \hline
%&  & &  &  & & \\ \hline
\end{tabular}
\raggedright $\mathcal{D}$: Domain type (source or target); $\mathcal{C}$: Number of classes; $\mathcal{L}$: Average sentence length; $\mathcal{N}$: Number of sentences; $\mathcal{V}$: Vocabu- lary size; $\mathcal{V}_{pre}$: Number of words found in the pre-trained word vectors.
\end{table}
One of the primary challenges in TL is to find a criterion for choosing a compatible source dataset for transferring knowledge to a given target dataset.
Once this source dataset has been identified, its corresponding DNN can be used for TL. Semantic similarities between the source and target databases can be useful in determining this criterion. 
%<SUGGESTED SENTENCE IN LIEU OF ABOVE: Given several DNNs trained respectively using separate source datasets, evolving  a criterion to select the best set of source datasets compatible to a given target dataset, constitutes one of the main challenges in TL.> 
TF\-IDF \cite{rajaraman_ullman_2011} and Doc2vec \cite{le14mikolov} have been used to calculate the semantic similarities between a source and target dataset. It may be noted that though these similarities have a statistical founding, the high computation time incurred and their inability to capture proper semantics render them ineffective. Given such scenarios, one of the objectives of the work described herein, is to segregate the available datasets into compatible sets of sources and targets by extensively analyzing the transfer behavior of the associated DNNs through empirical methods.
In the following sections, we will present a transfer method and the associated experiments carried out. 
%From the available open source datasets, Amazon (AMZ-2, AMZ-5) and Yelp reviews (YELP-2) becomes our obvious choice for the task of sentiment classification. The reason being is their large size and a huge vocabulary. After deciding on the source datasets, we chose the remaining small size datasets (MR, SST-2, SST-5, IMDb) as target domains for TL experiments.

\section{Hierarchical Transfer Method}
DNNs are composed of hidden layers arranged hierarchically one after another. More the number of hidden layers in a DNN, the deeper it gets. This also makes it bigger and complex. The hierarchical layered architecture provides flexibility and facilitates customization and thus, making it highly suitable for TL. One of the most straightforward ways of transferring knowledge is by initializing the parameters of a DNN to be trained on a target dataset to those of a DNN trained on a source dataset. This method is known to have produced successful results in various image processing applications. The same has been used herein for text classification.  We will first introduce the various layers and naming conventions used during the course of this work, followed by two different \textit{transfer settings} used in this paper. To avoid confusion, we  use the same nomenclature as in \cite{mou2016HowTA}. 

\begin{figure}[t]
	\centering
	\includegraphics[scale=.4]{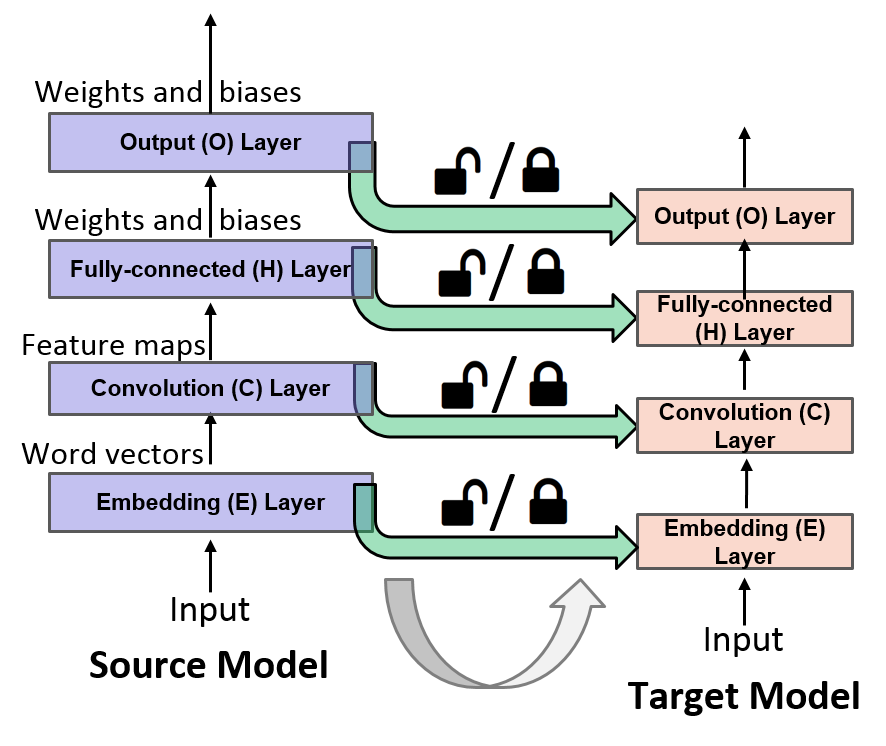}
	\caption{Hierarchical transfer of layers} 
	\label{fig:transfer}
\end{figure}

\subsection{Transferable Layers}
The various layers of a DNN can be abstracted as:
\begin{itemize}
\item \textbf{Embedding (E) layer:}
It is the very first layer of a DNN where the raw inputs are represented as blocks of much smaller units. For example, in the task of image classification, an image is represented by pixel values. Similarly, for text classification, a sentence is made up of words, where each word is represented as a \textit{d}-dimensional (D) vector ($d$ is an integer). This layer is found to learn features which are general across the different DNNs \cite{yosinski2014transferable}. 
%The embeddings form the inputs are learnable features which can be directly transferred from the source domain if present in target domain. For example, in the task of sentiment classification, the representation for the word \textit{awesome} could be transferable if found to be present in the target domain.

\item \textbf{Convolutional (C) layer:} It is an elemental layer of a Convolutional Neural Network (CNN) \cite{lecun1995convolutional}. The learnable features in a C layer are filters (or kernels) of much smaller size than the input. There could be multiple filters of different dimensions, which have a small receptive field and learn different features. For example, one feature could learn straight edges while another could learn the rounded edges in an input image. Similarly, for text inputs, a filter of size two learns bi-gram features. The weights of the filters form the transferable features of a C layer.

\item \textbf{Fully connected hidden (H) layer:} It is a regular hidden layer of a multi-layer perceptron which makes higher order decisions. It receives inputs from a preceding layer and performs a dot product with a weight matrix (a bias is also added occasionally) which are then passed through a non-linear activation function. The weights and biases form the transferable features.

\item \textbf{Output (O) layer:} The output layer is responsible for making the final decision which could be in the form of a continuous value (regression) or a class (classification) to which the input belongs. Similar to an H-layer, its weights and biases are transferable.
\end{itemize}

\subsection{Fine-tune or Freeze}
﻿After transferring the different layers (E, C, H and O), the parameters may be allowed to fine-tune if the labeled data is available in the target dataset.
The parameters could be left frozen if the labeled data is scarce. Freezing a layer means its parameters may not learn, i.e., the errors are not backpropagated.
Since there is a lesser number of parameters to learn, freezing a layer serves to reduce the size of the model. We performed transfer experiments under two settings - Fine-Tune (\faUnlock) and Frozen (\faLock). Two different settings allow for investigating into 
the transferability of the various layers within a DNN. Fig. \ref{fig:transfer} shows a hierarchical transfer between a Source Model and a Transfer Model. As can be seen from the Fig. \ref{fig:transfer}, the layers are transferred one after the another and are either allowed to fine-tune (\faUnlock) or left frozen (\faLock).

\begin{figure}[t]
	\centering
	\includegraphics[scale=.32]{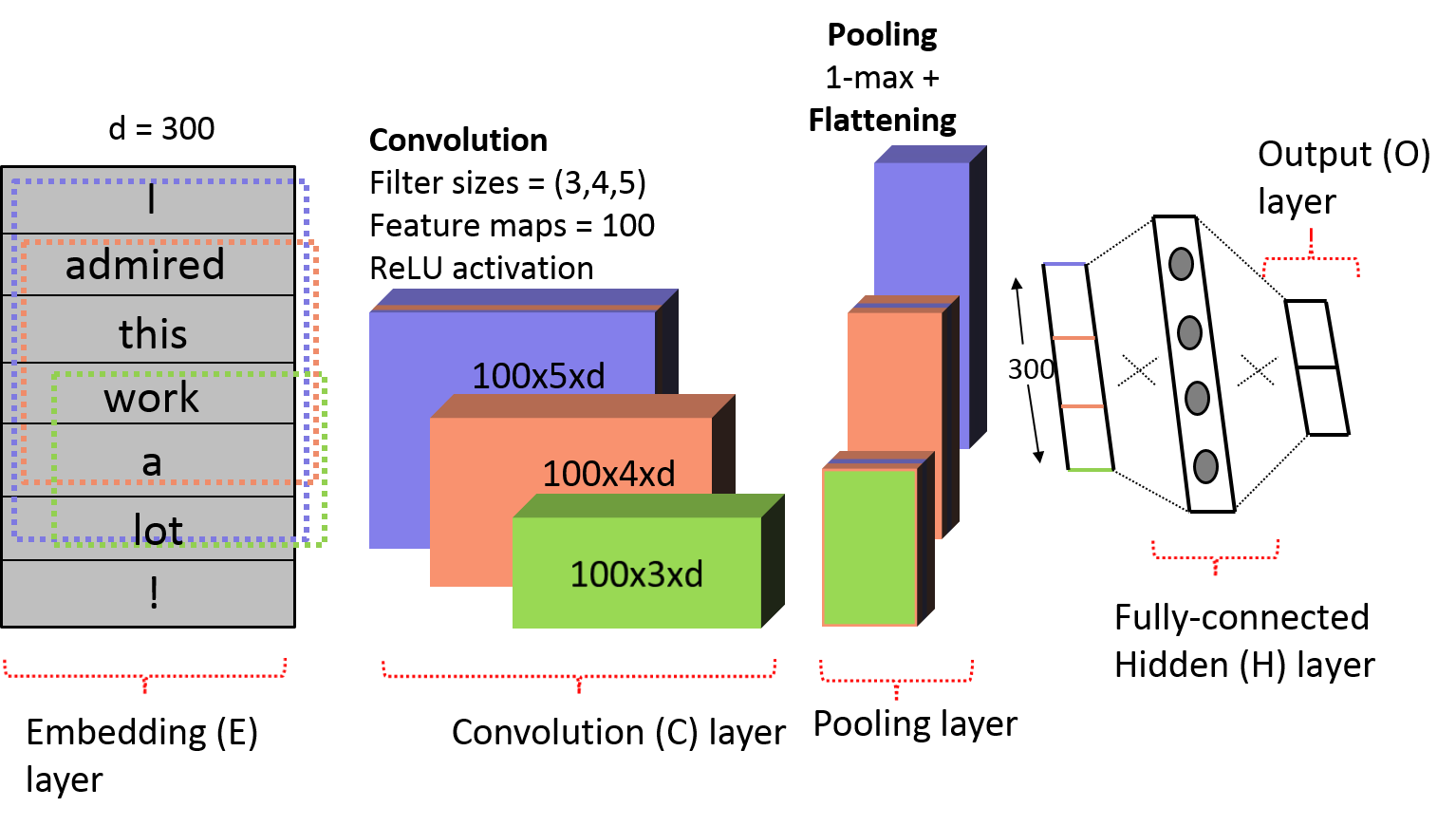}
	\caption{Architecture of the baseline CNN model} 
	\label{fig:arch}
\end{figure}

\begin{table}[tbp]
\footnotesize
\centering
\caption{Baseline parameters}
\label{tab2}
\begin{tabular}{l|l}
 Attribute& Value \\ \hline \hline
 Word embeddings& Google word2vec \\ \hline
 Filter region size& (3,4,5) \\ \hline
 Feature maps & 100 \\ \hline
 Activation function & ReLU \cite{maas13rectifiernonlinearities} \\ \hline
 Pooling & 1-max pooling \\ \hline
 Dropout rate & 0.6 \\ \hline
 \textit{l}2 norm constraint & 3 \\ \hline
 FC hidden layers & 1 \\ \hline
 FC activation & Iden \\ \hline
 
\end{tabular}
%\raggedright This is where authors provide additional information about the data, including whatever notes are needed.
\end{table}
\section{Transfer Experiments}
We first analyze the effect of TL for the task of text classification. 
In our experiments, we have adopted a slight variant of the CNN model (a fully connected hidden layer just after the convolution layer) proposed by \cite{kim2014convolutional}. The architecture of our CNN model is shown in Fig. \ref{fig:arch} while Table \ref{tab2} provides the decision parameters used in our CNN model.

\renewcommand{\thefootnote}{\faGithub}
To create a baseline, we initially trained our CNN model on all the datasets. Table \ref{tab3} provides a comparison of the accuracies  obtained using our CNN model along with those reported by others on the same datasets. The accuracies reported are based on test data if it is available for a  given dataset otherwise a 10-fold cross-validation (CV) is performed. For each fold, 10\% of the training data is randomly selected as the test data.  The optimization algorithm used is of the stochastic gradient descent type trained over shuffled mini-batches using the Adadelta update rule \cite{zeiler2012adadelta} and is similar to the one reported in \cite{kim2014convolutional} . Each of the experiment is repeated five times and the results reported are the average over all the repetitions.
Accuracy results reported in Table \ref{tab3} show that our baseline CNN model has satisfactory performance and potential for being used in TL experiments\footnote{\url{https://github.com/tushar-semwal/TransferLearning_CNN_TextClassification}}.

\begin{table}[t]
\footnotesize
\centering
\caption{Average accuracy (\%) of the baseline CNN model}
\label{tab3}
\begin{tabular}{c|c|l}
Dataset & Acc.(\%) & Similar model \\ \hline \hline
AMZ-5 & 58.1 & 59.5 (char-CNN \cite{zhang2015character}) \\ %\hline
AMZ-2 & 93.6 & 95.5 (VDCNN \cite{conneau2016verydeep})\\ %\hline
YELP-2 & 96.3 & 95.7 (VDCNN \cite{conneau2016verydeep}) \\ \hline
IMDb & 87.7 & 89.3 (Non-NN-Dong \cite{dong2015statistical}) \\ %\hline
SST-2 & 87.3 & 87.2 (CNN-Kim \cite{kim2014convolutional}) \\ %\hline
SST-5 & 47.2 & 48.5 (DCNN \cite{kalchbrenner2014convolutional}) \\ %\hline
MR & 81.8 &  81.5 (CNN-Kim \cite{kim2014convolutional})\\ \hline
%SUBJ & 93.2 & 93.4 (CNN-Kim)\\ \hline

\end{tabular}
%\raggedright This is where authors provide additional information about the data, including whatever notes are needed.
\end{table}

\section{Results}
Mou et al. \cite{mou2016HowTA} have discovered evidence which shows that TL in NLP applications is more sensitive to the semantics than in image processing. Hence, the first problem at hand is to find a semantically similar source dataset for a given target dataset. In order to verify and perform an in-depth analyses of the semantic relatedness, we chose sentiment classification as our first set of experiments. From the available datasets, AMZ and YELP were selected as the source datasets while MR, IMDb and SST were chosen to be the target datasets. Since IMDb is relatively a larger dataset than MR and SST, we also show results when IMDb acts as a source.  We chose three different kinds of datasets, all having two classes, with MR and SST-2 being the smaller sized datasets and IMDb as a moderately large sized dataset. To verify semantic dissimilarity, SST-5 was chosen as a 5-class dataset. We conducted transfer experiments for both - fine-tune (\faUnlock) and frozen (\faLock) settings for various $\mathcal{D}_s \twoheadrightarrow \mathcal{D}_t$ combinations. In the following subsections, we first study the semantic relatedness between a given task and the various source datasets, followed by a layer-by-layer analyses. We then show the effect of different hyper-parameter settings on the accuracy results after the transfer of layers.

\begin{table*}[tbp]
\tiny
\centering
\caption{Transfer accuracy (\%) of different $\mathcal{D}_s \twoheadrightarrow \mathcal{D}_t$ pairs}
\label{tab4}
\begin{tabular}{l||c|c|c|c||c|c|c|c||c|c||c|c}
 & \multicolumn{4}{c}{$\twoheadrightarrow$ MR } & \multicolumn{4}{c}{$\twoheadrightarrow$ SST-2}& \multicolumn{2}{c}{$\twoheadrightarrow$ SST-5}& \multicolumn{2}{c}{$\twoheadrightarrow$ IMDb}\\
    \hhline{~------------} 
%E*~C*~H*~O* & & \\ 
%E+~C*~H*~O* & & \\
\textbf{Setting} & AMZ-2 & AMZ-5 & YELP-2  & IMDb & AMZ-2 & AMZ-5 & YELP-2 & IMDb & AMZ-2 & AMZ-5 & AMZ-2 & YELP-2 \\ \hline \hline
E\faLock~\hspace{0.05cm}C\faAsterisk~H\faAsterisk~O\faAsterisk & 81.89 & 82.08 & 81.20 & 81.76 & 87.16 & 87.33 & 86.72 & 86.56 & 48.06 & 48.09 & 88.91 & 87.77 \\ 
  
E\faLock~\hspace{0.05cm}C\faLock~\hspace{0.05cm}H\faAsterisk~O\faAsterisk & 82.48 & 81.28 & 76.50 & 81.65 & 85.11 & 85.66 & 80.99 & 82.37 & 39.45 & 39.86 & 89.13 & 85.02 \\ 
 
E\faLock~\hspace{0.05cm}C\faLock~\hspace{0.05cm}H\faLock~\hspace{0.05cm}O\faAsterisk & 81.28 & 81.11 & 74.27 & 81.43 & 83.90 & 84.62 & 79.24 & 81.41 & 34.79 & 36.42 & 88.81 & 84.09 \\ 

E\faLock~\hspace{0.05cm}C\faLock~\hspace{0.05cm}H\faLock~\hspace{0.05cm}O\faLock & 79.64 & --- & 74.18 & 79.77 & 50.08 & --- & 77.70 & 49.91 & --- & 40.95 & 87.26 & 83.77 \\ 

E\faUnlock~C\faAsterisk~H\faAsterisk~O\faAsterisk & 82.22 & 82.29 & 81.30 & 82.26 & 88.08 & 87.62 & \textbf{87.66} & 87.16 & \textbf{48.70} & 48.58 & 88.67 & 88.08 \\ 

E\faUnlock~C\faUnlock~H\faAsterisk~O\faAsterisk & 83.73 & 83.12 & 82.00 & 83.08 & \textbf{89.38} & 89.03 & 86.89 & 87.46 & 48.46 & 48.50 & 89.81 & 88.24 \\ 

E\faUnlock~C\faUnlock~H\faUnlock~O\faAsterisk & \textbf{84.13} & \textbf{83.49} & \textbf{82.21} & \textbf{83.30} & 89.14 & \textbf{89.58} & 87.18 & 87.44 & 48.29 & 48.49 & \textbf{89.9}2 & \textbf{87.94} \\ 

E\faUnlock~C\faUnlock~H\faUnlock~O\faUnlock & 84.10 & --- & 81.47 & 83.22 & 89.30 & --- & 86.83 & \textbf{88.19} & --- & \textbf{48.68} & 89.91 & 87.92 \\ \hline

\end{tabular}
\raggedright ~ ~ ~~ ~~\faLock: Parameters are transferred and frozen; \faUnlock: Fine-tuning of transferred parameters is allowed; \faAsterisk: Parameters are randomly initialized and allowed to fine-tune; The AMZ-2, AMZ-5 and YELP-2 forms the source dataset ($\mathcal{D}_s$) while MR, SST-2 and SST-5 are the target datasets ($\mathcal{D}_t$). IMDb is in both $\mathcal{D}_s$ and $\mathcal{D}_t$. The top most column headings are the target datasets represented as $\twoheadrightarrow \mathcal{D}_t$ while the sub-column headings are $\mathcal{D}_s$.
\end{table*}

\subsection{Semantic relatedness}
Table \ref{tab4} reports the transfer accuracies for different $\mathcal{D}_s \twoheadrightarrow \mathcal{D}_t$ combinations wherein each column represent a $\mathcal{D}_s \twoheadrightarrow \mathcal{D}_t$ pair (for example, AMZ-2 $\twoheadrightarrow$ MR) and each row denotes a particular ECHO (for example, E\faUnlock ~C\faLock ~H\faLock ~O\faLock) setting. The overall results show that enabling the layers to fine-tune (\faUnlock) always results in accuracy gain. Thus, if the labeled data is available, it is the best practice to allow the learning of the transferred parameters. 
    
Unlike image processing tasks, a neural network is found to be transferable in NLP only for semantically similar datasets . Though our findings are in favor of the reported conclusion, we were curious to know the extent of semantic similarity in the task of text classification. Hence, we conducted transfer experiments for a given sentence classification task by transferring knowledge obtained from DNNs trained using different source datasets. The following section discusses experimental results on different target datasets.

\begin{table}[t]
\footnotesize
\centering
\caption{Words not present in the source dataset}
\label{tab5}
\begin{tabular}{c||c|c|c|c}
%\hline
 Dataset & MR  & SST-2 & SST-5 & IMDb \\ \hline \hline
AMZ-2 &  509 & 2477 & 331 & 8273 \\ 
AMZ-5 &  548 & 2492  & 348 & 8393  \\ 
YELP-2 & 2216 & 3720 & 1697& 21086 \\ 
IMDb &  2723  & 4003 & 2058 & --- \\ \hline
%&  & &  &  & & \\ \hline
\end{tabular}
%\raggedright This is where authors provide additional information about the data, including whatever notes are needed.
\end{table}

\subsubsection{MR} In the task of sentiment classification, IMDb (movie reviews) is  semantically more similar to MR (movie reviews) than AMZ (product reviews) and YELP (business reviews) datasets. However, the accuracy results say otherwise. The best AMZ-2 $\twoheadrightarrow$ MR accuracy (84.13\%) is greater than IMDb $\twoheadrightarrow$ MR (83.30\%) pair. This is because of the large size of the source dataset AMZ-2 and the richer embeddings therein which have been able to learn better contextual information than the IMDb dataset. This reasoning can again be justified with the case of AMZ-5 $\twoheadrightarrow$ MR transfer wherein though the two tasks have unequal number of classes to predict, AMZ-5 provides a better or at least comparable accuracy with the other source datasets (AMZ-2, YELP-2 and IMDb) due to its semantic similarity and a larger vocabulary.

\subsubsection{SST-2}Since SST-2 is an extension of the MR dataset, both are semantically similar, and thus the transfer results obtained with the same source datasets were found to be comparable and follow a similar pattern.

\renewcommand{\thefootnote}{\faGears}
\subsubsection{SST-5} SST-5 is a 5-class fine-grained version of the SST dataset. In order to evaluate the impact of the number of classes on the transfer accuracy, two experiments, AMZ-2 $\twoheadrightarrow$ SST-5 and  AMZ-5 $\twoheadrightarrow$ SST-5, were carried out. As can be seen from the Table \ref{tab4}, the accuracies obtained after transferring only the first layer (E\faLock~C\faAsterisk~H\faAsterisk~O\faAsterisk~and E\faUnlock~C\faAsterisk~H\faAsterisk~O\faAsterisk) are same for both of the source datasets. However, as we transfer further layers, the drop in the accuracy for AMZ-2 transfer is higher than that for AMZ-5. Thus, even if the two datasets (AMZ-2 and AMZ-5\footnote{AMZ-2 has \{0,1\} while AMZ-5 has \{0,1,2,3,4\} as the class labels, i.e., AMZ-5 is a fine-grained version.}) contain similar sentences, the transfer accuracy may decrease for the C-layer and beyond. This is known to be a \textit{negative} transfer. It should be noted that for this case, the negative transfer is visible only for the \faLock~ setting which shows the actual transfer capabilities of the source dataset. One may thus infer that the semantic similarity between the source and target dataset for the Task-2 is low. For \faUnlock~setting, the DNN is allowed to fine-tune its parameters thereby bringing the accuracy to the same level.

\subsubsection{IMDb}  We conducted two experiments with AMZ-2 and YELP-2 as the source datasets. As can be seen, the best accuracy for AMZ-2 $\twoheadrightarrow$ IMDb (89.92\%) is higher than YELP-2 $\twoheadrightarrow$ IMDb (88.24\%). From these experiments, we found that even though YELP-2 is a large and similar dataset, it produces a lower transfer gain in terms of accuracy as compared to AMZ-2. Table \ref{tab5} reports the Out-Of-Vocabulary (OOV) words for each target dataset not available in the source dataset. As can be seen from the Table \ref{tab5}, from the total of 81,321 words, 21,086 were not present in the YELP dataset. This shows that OOV could be one of the factors before deciding on the source dataset.

\subsection{Transferability of layers}

As pointed out in \cite{yosinski2014transferable}, the generality of the layers of a DNN decreases as we move towards the output decision layer. The layers close to the output layer, learn features which are more task specific while the layers far away learn general features. Even though the results shown in Table \ref{tab4} depict a similar trend in the accuracy values \cite{yosinski2014transferable,mou2016HowTA}, none of the previous research provides concrete conclusions on positive and negative transfer scenarios. In this subsection, we present a layer-by-layer analyses from the extensive experiments performed on different $\mathcal{D}_s~\twoheadrightarrow~\mathcal{D}_t$ pairs.

\subsubsection{E-layer} As can be seen in Table \ref{tab4}, transferring the E-layer in both the settings (E\faLock~C\faAsterisk~H\faAsterisk~O\faAsterisk~and E\faUnlock~C\faAsterisk~H\faAsterisk~O\faAsterisk), always results in a positive transfer, independent of the difference in the number of classes between the source and target datasets. However, the amount of improvement for a semantically similar task is found to be dependent on the OOV metric, as shown in Table \ref{tab5}. For example, in cases of YELP-2 $\twoheadrightarrow$ MR and IMDb $\twoheadrightarrow$ MR,  the OOV words in YELP-2 and IMDb are fourfold more than that in AMZ-2. One may conclude that the OOV metric could be a significant criterion for deciding the source dataset in scenarios involving the transfer of the E-layer. A plausible explanation is that an E-layer represent words as vectors with inherent context information. Hence, greater the overlap in the vocabulary of the source and target datasets, greater the transfer of the context information and thus, improving the classification accuracy.

\subsubsection{C-layer}  For the \faLock~setting, other than AMZ-2 $\twoheadrightarrow$ MR and AMZ-2 $\twoheadrightarrow$ IMDb, none of the $\mathcal{D}_s~\twoheadrightarrow~\mathcal{D}_t$ pairs produced a positive transfer after the transfer of the C-layer. In contrast, transfer of the C-layer under the \faUnlock~setting caused an improvement in the accuracy for the majority of dataset pairs. Even though Mou et al. \cite{mou2016HowTA} have thrown some light on this phenomenon, the actual behavior of the C-layer in the transfer experiments is still unclear.  

The Convolution layer is composed of different filter maps which learn the various kinds of features present in the input. Depending upon the application, the input could be a 2-D image or a 1-D sentence. Due to spatial variance found in the images, the features learned by the filter maps are bound to be common with another dataset. For instance, the horizontal and vertical edges have a high chance of being present in images from another dataset. Thus, this forms a significant reason behind the successful transfer of C-layers in image-based applications. Contrary to this, sentences are 1-D and presented in the shape of 2-D inputs, with the height equal to the number of words or characters and the width as the dimension of the word vector representation. The filters are convolved along a single axis (height) with each filter of different sizes capturing different contextual information for each word (or character) in a given sentence. Hence, it is more likely that the datasets with similar semantics will not have the same contextual information. This explains the transfer results under \faLock~setting.

\subsubsection{H-layer} 
As can be seen in Table \ref{tab4}, under \faLock~setting,  the transfer of the H-layer decreases the performance for all the $\mathcal{D}_s~\twoheadrightarrow~\mathcal{D}_t$ pairs. However, for \faUnlock~setting, E\faUnlock~C\faUnlock~H\faUnlock~O\faAsterisk~emerged to be the best setting in majority  of the transfer experiments. An H-layer learns top level features (for example, sentence vectors) which could be similar for semantically related datasets. Thus, even though the H-layer is very close to the output decision layer it is capable of producing accuracy gains when transferred.

\subsubsection{O-layer} 
Under both the settings (\faLock~and \faUnlock), transfer of the output layer always impeded the accuracy values. This suggests that this layer is specific to a particular task or dataset and thus is best avoided in the transfer.

\subsection{Effect of Parameters}
After analyzing the semantic relatedness and transferability of each layer, we now consider the effect of the different architecture parameters for both the source and target datasets on the transfer accuracy. In general, training a large source dataset is a costly affair. Thus, in most of the scenarios, the available source model for TL is trained on a particular parameter configuration. For a given experiment, only a desired parameter is varied while all others (refer Table \ref{tab2}) are kept constant. In this paper, we present the results of transfer experiments for only the most significant set of parameters. For every parameter we analyze, we repeated each experiment five times, where each replication is a run of 10-fold CV. The values reported are the average over all the repetitions.

\begin{table}[t]
\footnotesize
\centering
\caption{Effect of activation function on the transfer accuracy}
\label{tab6}
\begin{tabular}{c||c|c|c}
%\hline
 \textbf{Model} & MR-Iden  & MR-Tanh & MR-ReLU \\ \hline 
AMZ-2-Iden &  \textbf{81.28},\textbf{84.13} & \textbf{81.11},\textbf{84.13}
 & 81.08,\textbf{84.25}  \\ 
AMZ-2-Tanh &  81.14,83.71& 81.05,83.94 &  \textbf{81.11},83.94 \\ 
AMZ-2-ReLU & 80.97,83.90 & 80.99,83.90& 81.01,83.81\\ 
\hline

%&  & &  &  & & \\ \hline
\end{tabular}
%\raggedright This is where authors provide additional information about the data, including whatever notes are needed.
\end{table}
\subsubsection{Activation function}
In order to capture the relation between the input vectors and output labels, activation functions are used. These are either linear or nonlinear depending upon the input data. We chose one liner and two nonlinear activation functions which are widely used by the machine learning practitioners. The activation function is applied to the output of both the C- and H-layers. We have performed the experiments with Rectified Linear Units (ReLU) \cite{maas13rectifiernonlinearities}, hyperbolic tangent (Tanh) and Identity (Iden) functions. Table \ref{tab6} reports the results (each cell is a \faLock,\faUnlock~pair) for different activation functions used on both the source and target datasets. Iden is a linear activation function whose output is equal to the input.

Though \faLock~setting yields a lower accuracy value, from an experimental point of view, it provides better insights into the transfer properties of the source dataset. When no activation has been used in the source dataset (AMZ-2-Iden), the MR-Iden under the \faLock~setting shows a slightly better result (81.28\%) than the other two functions. The results may be evident as both the models (source and target datasets) use the same activation function. However, if fine-tuning (\faUnlock) is allowed, MR-ReLU provides the best-reported accuracy (84.25\%) which is in agreement with the results reported in \cite{zhang2015sensitivity}. 
While we achieved state-of-the-art results with the source model trained without using any activation function, an interesting point to perceive was that with other activation functions, the transfer results were found to be inferior. This suggests that it is preferable to choose a source model trained without using any nonlinear activation function.

\begin{table}[t]
\footnotesize
\centering
\caption{Effect of dropout rate on the transfer accuracy}
\label{tab7}
\begin{tabular}{c||c|c|c||c}
%\hline
& \multicolumn{3}{c}{AMZ-2 $\twoheadrightarrow$  } & \multicolumn{1}{c}{AMZ-5 $\twoheadrightarrow$ }\\
    \hhline{~----}
\textbf{Dropout}  & MR  & SST-2 &  IMDb & SST-5 \\ \hline 
0.0 & 83.16 & 86.67 & 89.87 & 47.58 \\ 
0.1 & 83.14 & 87.47 & 90.02 & 47.84  \\ 
0.2 & 83.43 & 87.86 & 89.89 & \textbf{49.32} \\ 
0.3 & 83.42 & 88.75 & 89.78 & 48.90 \\ 
0.4 & 83.84 & 88.93 & 89.94 & 47.94\\ 
0.5 & 84.04 & 89.01 & \textbf{90.05} & 48.75\\ 
0.6 & 84.13 & \textbf{89.38} & 89.94 & 48.40\\ 
0.7 & \textbf{84.35} & 89.16 & 89.84 & 48.08\\ 
0.8 & 83.68 & 88.06 & 89.66 & 46.69\\ 
0.9 & 83.73 & 87.62 & 89.08 & 42.59\\ 

\hline

%&  & &  &  & & \\ \hline
\end{tabular}
%\raggedright This is where authors provide additional information about the data, including whatever notes are needed.
\end{table}

\subsubsection{Regularization}
Dropout \cite{srivastava2014dropout} is found to be a good regularizer in text classification \cite{zhang2015sensitivity,kim2014convolutional}. An extensive analysis on CNN by Zhang et al. \cite{zhang2015sensitivity} reveals that a change in the dropout rate has little effect on the accuracy values while a significant dropout may adversely affect the model performance. However, this may not be the case in TL scenarios. An intuitive reason could be that the knowledge accumulated within the source model may over-fit the target model and thus a high dropout becomes necessary. This foreknowledge is in contrast to the results portrayed in \cite{zhang2015sensitivity} where low dropouts values are preferred to work better. Thus, we were curious to know the range of dropout rates which can aid in achieving better transfer models.

As per our baseline architecture (refer Table \ref{tab2}), we fixed the \textit{l}2 norm (another regularization technique) constraint to 3, while the dropout rate was varied from 0.0 to 0.9. We performed the experiments on different $\mathcal{D}_s$ $\twoheadrightarrow$ $\mathcal{D}_t$ pairs which have shown positive transfer for a given transfer setting. Table \ref{tab7} depicts the accuracies achieved under different dropout rates. Each $\mathcal{D}_s$ $\twoheadrightarrow$ $\mathcal{D}_t$ pair is arranged in the decreasing order of their transfer performance, i.e., AMZ-2 $\twoheadrightarrow$ MR has the best while AMZ-5 $\twoheadrightarrow$ SST-5 has the least 

From Table \ref{tab7}, it can be seen that a high dropout is required for a $\mathcal{D}_s$ $\twoheadrightarrow$ $\mathcal{D}_t$ pair delivering high transfer performance. In contrast, dropout rate falls with the decrease in the transfer potential between the source and target datasets. This means that as the transfer performance decreases, the model behaves similar to the case when it is trained without the transfer. Thus, the appropriate values for dropout rates fall in the range varying from 0.0 to 0.5 as suggested in \cite{zhang2015sensitivity}.

\begin{table}[t]
\footnotesize
\centering
\caption{A comparison of accuracy and number of parameters against other methods }
\label{tab8}
\begin{tabular}{l||c|c|c|l}
%\hline
 Method & MR  & SST-2 & SST-5 & \#Parameters \\ \hline \hline
CNN-Kim \cite{kim2014convolutional} &  81.5	 & 87.2 & 48 & $\approx$360K \\ 
Dep-CNN \cite{ma2015dependency} &  81.9 & --- & 49.5 & $\approx$840K  \\ 
DSCNN-P \cite{zhang2016dependency} & 82.2 & 89.1 & \textbf{52.6} & $\approx$1488K \\ 
TE-LSTM+c \cite{huang2017te_lstm} & 81.6 & \textbf{89.4} & 52.3 & $\approx$919K \\ 
\hline
E\faAsterisk~C\faAsterisk~H\faAsterisk~O\faAsterisk & 87.20 & 87.1 & 48 & $\approx$390K\\
E\faLock~\hspace{0.07cm}C\faAsterisk~H\faAsterisk~O\faAsterisk & 81.8 & 87.1 & 48.09 & $\approx$390K \\ 
  
E\faLock~\hspace{0.07cm}C\faLock~\hspace{0.08cm}H\faAsterisk~O\faAsterisk & 82.4 & 85.1 & 39.8 & $\approx$30K  \\ 
 
E\faLock~\hspace{0.07cm}C\faLock~\hspace{0.09cm}H\faLock~\hspace{0.07cm}O\faAsterisk & 81.2 & 83.9 & 36.4 & $\approx$0\\ 

E\faLock~\hspace{0.07cm}C\faLock~\hspace{0.09cm}H\faLock~\hspace{0.07cm}O\faLock & 79.6 & 50.0 & 40.9 & $\approx$0\\ 

E\faUnlock~C\faAsterisk~H\faAsterisk~O\faAsterisk & 82.22 & 88.0 & 48.5 & $\approx$390K\\ 

E\faUnlock~C\faUnlock~H\faAsterisk~O\faAsterisk & 83.73 & \textbf{89.3} & 48.5 & $\approx$390K\\ 

E\faUnlock~C\faUnlock~H\faUnlock~O\faAsterisk & \textbf{84.13} & 89.1 & 48.4 & $\approx$390K \\ 

E\faUnlock~C\faUnlock~H\faUnlock~O\faUnlock & 84.10 & 89.3 & \textbf{48.6} & $\approx$390K \\ 
\hline

\hline
%&  & &  &  & & \\ \hline
\end{tabular}
\raggedright \textbf{CNN-Kim}: CNN for Sentiment Classification by Kim (2014) \cite{kim2014convolutional}. \textbf{Dep-CNN}: Dependency based CNN for sentence embedd- ing by Ma et al. (2015) \cite{ma2015dependency}. \textbf{DSCNN-P}: Dependency Sensitive CNN with Pre-trained encoders for sentence modeling by Zhang et al. (2016) \cite{zhang2016dependency}. \textbf{TE-LSTM+c}: Part-Of-Speech (POS) Tag Embedded (TE) Long Short Term Memory (LSTM) model Combined (+c) with word representations by Huang et al. (2017) \cite{huang2017te_lstm}.
\end{table}

\subsection{Comparison with state-of-the-art results}
Having presented the semantic relatedness and effect of various hyperparameters on the transfer performance, we now compare our results with some of the near state-of-the-art models. Table \ref{tab8} reports the accuracy values and the number of parameters (model size) for our TL augmented CNN model against different DNN models. As can be seen, even with a small size, our CNN model under E\faUnlock~C\faUnlock~H\faUnlock~O\faAsterisk~setting outperforms state-of-the-art results on the MR dataset. For SST-2 it shows comparable performance even though the size is less than half of the other methods (DSCNN-P and TE-LSTM+c). Though the results for SST-5 are poor, there is still an improvement in the accuracy after transferring the knowledge.  It may be noted that the number of parameters reported in Table \ref{tab8} do not include word embeddings  as they are same for other considered models. For a similar reason, the weights and the bias values of the O-layer are also not included.

\section{Conclusions}
In this paper, we have performed extensive empirical investigations on TL for Text Classification. We conducted transfer experiments on seven datasets out of which three were chosen as the source and remaining as the target datasets. Along with semantic analysis and transferability of different neural layers, we also study the effect of various hyperparameters on the transfer performance. In addition, we compared the accuracy value and model size of out TL augmented CNN model with other state-of-the-art methods. We present below our findings and the recommendations derived from them. 

\subsection{Summary of the main results}
%Though there have been several use cases of TL, it is rare to find work where the applied techniques and hyperparameters settings that can lead to successful transfer results, have been discussed. 
Some of our salient observations include:
\begin{itemize}
\item According to \cite{mou2016HowTA}, the transfer performance in NLP applications is prone to issues due to semantics. However, they do not present the extent of semantic matching which can aid in choosing a better source dataset. Through extensive experiments on different $\mathcal{D}_s \twoheadrightarrow \mathcal{D}_t$ pairs, we found that even a lesser semantically similar dataset can produce better transfer results than a highly similar dataset. For instance, even though IMDb is a movie review dataset, transfer performance for AMZ-2 $\twoheadrightarrow $ MR is greater than IMDb$ \twoheadrightarrow $MR, even though both IMDb and MR are movie review datasets while AMZ-2 is a business review dataset. 

\item Along with the semantics, transfer performance also depends on the size and OOV metric of the source dataset. %For example, AMZ-2 $\twoheadrightarrow $ IMDb produces better transfer results than YELP-2$\twoheadrightarrow $ IMDb 

\item Transferring the E-layer (for both \faLock~and \faUnlock~setting) always results in a positive transfer. However, the amount of accuracy gain depends on the OOV metric and the semantic matching between the source and target datasets.

\item Under \faLock~setting, it is less likely that transferring of C-layer and H-layer will result in transfer gains unless the source dataset is (i) semantically similar, (ii) has a low OOV and (iii) a large vocabulary size. In contrast, \faUnlock~produces appreciable transfer performance for the majority of the $\mathcal{D}_s \twoheadrightarrow \mathcal{D}_t$ pairs.

\item Irrespective of the transfer settings, the transfer of the O-layer  yields  negligible or no transfer gains. This is akin to what is reported in \cite{yosinski2014transferable,mou2016HowTA}.

\item Intuitively one may conclude that the transfer performance will be higher for the case when the DNN models for both the source and target datasets have the same activation function. Nevertheless, our reported results do not agree to this entirely. If a source model is trained with a non-linear activation function, then the choice of activation function for the target dataset will depend on the target dataset and needs to be found empirically. However, if a source model does not have any activation function (AMZ-2-Iden), then the results follow from the parameter analysis by Zhang et al. \cite{zhang2015sensitivity}. For example, with AMZ-2-Iden as the source model, MR-ReLU provides the best transfer accuracy, similar to the results presented in \cite{zhang2015sensitivity}.

\item A significant dropout (0.5 to 0.7) may be required if the source dataset is highly suitable for a target dataset. A smaller dropout rate delivers better performance for $\mathcal{D}_s \twoheadrightarrow \mathcal{D}_t$ pairs having low transfer potential.
\end{itemize}

\subsection{Suggested best practices}
\begin{itemize}
\item If available, start by choosing a source dataset, which has richer embeddings with low OOV metric. Prefer the source dataset having a large vocabulary size and which is partially similar to the target dataset. Choosing a smaller source dataset having a high semantic similarity to the target dataset may not yield desired results.

\item If possible, always transfer the E-layer for a chosen source model. 

\item In general, consider the transfer of the C- and H-layers under the \faUnlock~setting. One can consider freezing these layers only if they need a reduction in the model size and are sure about the semantic matching between the source and target datasets. Do not consider the transfer of the H-layer if the number of output classes is different for source and target datasets.

\item Never transfer the O-layer under the \faLock~setting. However, one can consider  transferring the O-layer provided the source and target datasets are highly similar, and fine-tuning of the parameters is allowed while training the model.

\item Prefer a source model which is trained with no activation function. Else, if such a  model is not available, consider referring to \cite{zhang2015sensitivity} to decide the initial parameters.

\item Use a dropout rate in the range varying from 0.4 to 0.7 for highly matching datasets. Consider decreasing the dropout value when the transfer potential is on par with an average performing model.
\end{itemize}

In the future, we would like to analyze the transfer performance for different NLP tasks such as Named-Entity Recognition, Parts-of-Speech tagging, machine translation and dialogue generation using different DNN models such as a Recursive Neural Network.

\end{document}